\documentclass{article} 
\usepackage{iclr2026_conference,times}
\usepackage{graphicx}
\usepackage{amsmath}
\usepackage{amssymb}
\usepackage{booktabs}

\usepackage{amsfonts}       
\usepackage{nicefrac}       
\usepackage{microtype}      

\usepackage{makecell}
\usepackage{amsthm}
\usepackage{bm}
\usepackage{multicol}
\usepackage{colortbl}
\usepackage{multirow}
\usepackage{enumitem}
\usepackage{bbding}
\usepackage{color}

\usepackage{algorithm}
\usepackage{listings}
\usepackage{natbib}

\usepackage{etoolbox}

\usepackage{wrapfig} 

\usepackage{makecell}
\usepackage{amsthm}
\usepackage{bm}
\usepackage{multicol}
\usepackage{colortbl}
\usepackage{amsmath}
\usepackage{multirow}
\usepackage{graphicx}
\usepackage{subfigure}
\usepackage{enumitem}
\usepackage{bbding}
\usepackage{color}

\usepackage{algorithm}
\usepackage{listings}

\usepackage{etoolbox}
\makeatletter
\AfterEndEnvironment{algorithm}{\let\@algcomment\relax}
\AtEndEnvironment{algorithm}{\kern2pt\hrule\relax\vskip3pt\@algcomment}
\let\@algcomment\relax
\newcommand\algcomment[1]{\def\@algcomment{\footnotesize#1}}
\renewcommand\fs@ruled{\def\@fs@cfont{\bfseries}\let\@fs@capt\floatc@ruled
  \def\@fs@pre{\hrule height.8pt depth0pt \kern2pt}%
  \def\@fs@post{}%
  \def\@fs@mid{\kern2pt\hrule\kern2pt}%
  \let\@fs@iftopcapt\iftrue}
\makeatother
\newcommand{\revise}[1]{{#1}}
\newcommand{\condenser}{\textsc{Condenser}}
\newcommand{\modelname}{{PromptHub}}
\newcommand{\Number}{{N}}
\newcommand{\numindex}{{n}}
\newcommand{\hubicon}{\raisebox{-0.1em}{\includegraphics[width=1em]{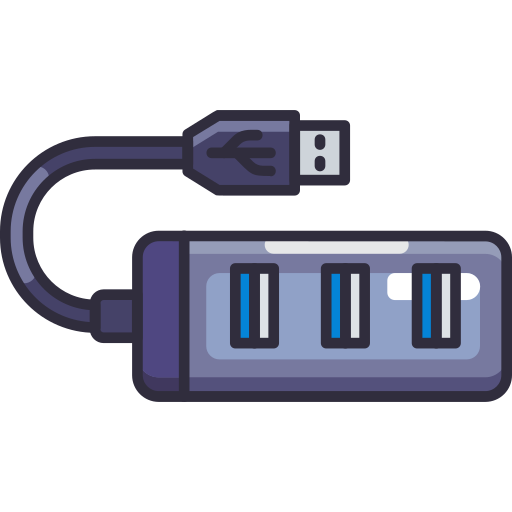}}\hspace{-0.1em}}

\usepackage{amsmath,amsfonts,bm}

\def\eqref#1{equation~\ref{#1}}

\def\1{\bm{1}}

\DeclareMathAlphabet{\mathsfit}{\encodingdefault}{\sfdefault}{m}{sl}
\SetMathAlphabet{\mathsfit}{bold}{\encodingdefault}{\sfdefault}{bx}{n}

\definecolor{ForestGreen}{RGB}{34,139,34}

\renewcommand{\paragraph}[1]{\medskip\noindent\textbf{#1.~}}

\usepackage{hyperref}
\usepackage{url}
\usepackage{cleveref}

\title{{\hubicon{} \modelname{}: Enhancing Multi-Prompt Visual In-Context Learning with Locality-Aware \mbox{Fusion}, Concentration and Alignment}}

\author{Tianci Luo$^{1,}\thanks{Equal contribution.}$\ \,, Jinpeng Wang$^{2, *,}\thanks{Corresponding authors.}$\ \,, Shiyu Qin$^{1}$, Niu Lian$^{2}$, \\ 
\textbf{Yan Feng$^{3}$, Bin Chen$^{2,\dagger}$, Chun Yuan$^{1,\dagger}$, Shu-Tao Xia$^{1}$}  \\
$^1$Tsinghua Shenzhen International Graduate School, Tsinghua University\\
$^2$Harbin Institute of Technology, Shenzhen\\
$^3$Meituan, Beijing \\
\texttt{ltc25@mails.tsinghua.edu.cn; wangjp26@gmail.com;}\\ \texttt{chenbin2021@hit.edu.cn;
yuanc@sz.tsinghua.edu.cn}
}

\iclrfinalcopy 
\begin{document}

\maketitle

\begin{abstract}

Visual In-Context Learning (VICL) aims to complete vision tasks by imitating pixel demonstrations. 
Recent work \citep{condenser} pioneered prompt fusion that combines the advantages of various demonstrations, which shows a promising way to extend VICL. 
Unfortunately, the patch-wise fusion framework and model-agnostic supervision hinder the exploitation of informative cues, thereby limiting performance gains. 
To overcome this deficiency, we introduce PromptHub, a framework that holistically strengthens multi-prompting through locality-aware fusion, concentration and alignment. 
PromptHub exploits spatial priors to capture richer contextual information, employs complementary concentration, alignment, and prediction objectives to mutually guide training, and incorporates data augmentation to further reinforce supervision.
Extensive experiments on three fundamental vision tasks demonstrate the superiority of PromptHub. Moreover, we validate its universality, transferability, and robustness across out-of-distribution settings, and various retrieval scenarios.
This work establishes a reliable locality-aware paradigm for prompt fusion, moving beyond prior patch-wise approaches.
Code is available at \url{https://github.com/luotc-why/ICLR26-PromptHub}.
\end{abstract}
\section{Introduction}
\label{sec: introduction}

Foundation models like GPT \citep{brown2020language}, Llama \citep{touvron2023llama}, Gemini \citep{team2023gemini} and Flamingo \citep{alayrac2022flamingo} have demonstrated the emerging ability of demonstration-based prompt learning, aka In-Context Learning (ICL) \citep{ICL_Survey_2024,zheng2023can,yang2023exploring}, which further facilitates their versatility in various tasks. 
The basic idea of ICL \citep{hendel2023context,wei2023larger,wei2022emergent,jiang2024dg} is to prompt models with some demonstrative input-output pairs in addition to the query input, which can enhance the answer robustness. The reliablity of ICL has been thoroughly validated  \citep{why_work_1, why_work_2, why_work_3, why_work_4}.
Recently, Visual ICL (VICL) \citep{bar2022visual,wang2023seggpt} have also become a popular topic, where pixel-space in-painting is the native paradigm. 

\begin{figure}[!t]
    \centering
    \includegraphics[width=0.95\columnwidth]{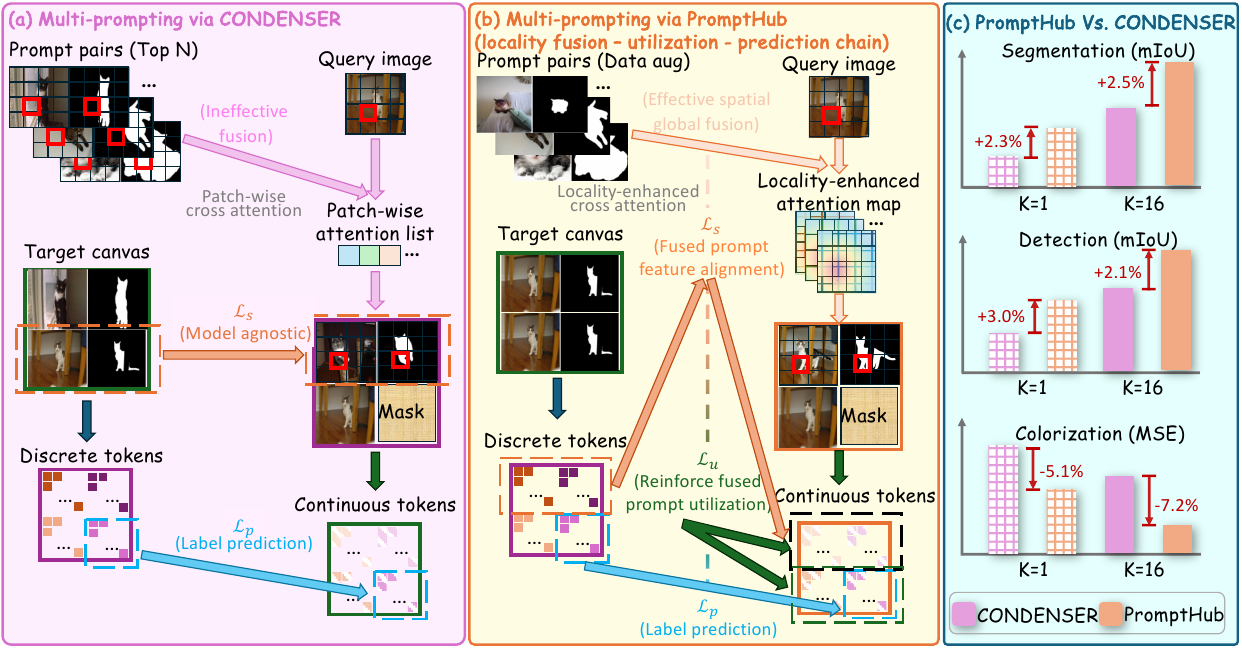}
    \caption{
    (a) \condenser{} performs patch-wise fusion to fuse composite prompt, while leveraging model-agnostic supervision signals at the input level.
    (b) PromptHub transcends \condenser{} by enforcing a locality-aware chain that unifies fusion-utilization-prediction. It aligns spatial priors into coherent prompt representations, reinforces the backbone’s concentration on fused cues, and integrates label prediction to maintain the integrity of VICL pipeline.
    (c) Comparison of \condenser{} and PromptHub across three tasks under both single-prompting and multi-prompting configurations.    
    }
    \vspace{-1em}
    \label{fig:intro}
\end{figure}

Choosing appropriate prompts is critical in VICL. 
Many recent works \citep{zhang2024makes,sun2023exploring,xu2024towards,hahaha} focused on optimizing retrievers to select better suited prompts, and \citet{zhang2024instruct} incorporated visual prompt tuning to enhance the robustness of VICL. 
Nature Language Processing (NLP) literature \citep{knnprompting,gao2024unifying} suggests multiple prompts can enhance ICL with mitigated bias and richer context, offering insights essential for advancing VICL.
Yet visual backbones like MAE-VQGAN typically restrict inputs to a single prompt, rendering multi-prompting non-trivial.
In practice, there are two heuristic strategies for extending single-prompting to multi-prompting, namely downscaling \citep{wang2023images} and ensemble \citep{sun2023exploring}. 
Building upon this, \condenser{} \citep{condenser} was the first to adopt prompt fusion \citep{pb}, integrating useful information from multiple prompts into a fused prompt, as illustrated in \Cref{fig:intro}(a).
However, its patch-wise fusion strategy results in substantial underuse of valuable cues, while model-agnostic supervision remains insufficient. 
Moreover, discrepancies between fused prompt and query pair may compels backbone to distrust fused representation, falling back on its own capacity for inference. This is precisely the situation we aim to avoid.

We break these limits by proposing \modelname{}, which aims at \textbf{(i)} integrate precise knowledge from diverse prompts, \textbf{(ii)} mitigate fused prompt's discrepancies to encourage the backbone’s effective trust and reliance on it, and \textbf{(iii)} ultimately yield superior VICL predictions. To achieve this, we propose a locality-aware fusion framework together with three cooperative learning objectives, as illustrated in \Cref{fig:intro}(b). Specifically, we introduce a locality prior that applies spatially decaying weights radiating from the current patch, thereby enhancing accurate feature extraction. This design allows the fusion process to retain a global receptive field while alleviating the adverse effects of border noise.
During the optimization of \modelname{}, we design three complementary objectives: 
\textbf{(i)} an end-to-end semantic integrity loss
to promote high-quality prompt fusion by aligning fused exemplars with query semantics, as semantically closer prompt generally benefit VICL; \textbf{(ii)} a utilization loss that mitigates discrepancies between the fused prompt and the query pair, thereby promoting the backbone's trust and reliance on the fused representation for imitation learning; and \textbf{(iii)} a label prediction loss, retained from \condenser{}, which serves as the base supervision to preserve VICL's contextual prediction behavior.
Additionally, we preliminarily explored VICL-oriented data augmentation strategies to enhance the robustness of \modelname{}. 
These designs realize chain-wide enhancements for VICL.

We evaluate \modelname{} on 
segmentation, detection, and colorization. 
As shown in \Cref{fig:intro}(c), extensive experiments demonstrated its superiority to state-of-the-art baselines.
We also demonstrate \modelname{}'s promising resource efficiency, transferability and robustness to various prompt retrieval strategies including random selection. 
Comprehensive ablations on diverse learning objectives, data augmentation techniques, and locality fusion, confirming the success of our paradigm design.
We further visualize the fused prompts, which validates the reliablity of PromptHub.
These findings strongly support the efficacy of \modelname{}, highlighting the significance of our approach in VICL. 

To sum up, we make the following contributions.
\setlist{itemsep=0.05em, topsep=0.05em}
\begin{itemize}[leftmargin=1.5em]

\item We introduce a locality-enhanced fusion strategy that balances spatial locality and receptive field, enabling more comprehensive extraction of effective information.

\item We propose three complementary learning objectives that collaboratively enhance prompt fusion quality, strengthen prompt concentration, and improve contextual prediction, further reinforced with VICL-specific data augmentation.

\item Extensive experiments show \modelname{}'s efficacy beyond state-of-the-art techniques. Promising results also suggest that it is transferable across domains and robust to prompt retrieval, establishing a reliable competitive new solution in VICL.
\end{itemize}

\section{Related Works}
\label{sec:related_work}

\subsection{Large Vision Models}

The field of computer vision has witnessed substantial advancements, driven by abundant foundational models \citep{chang2022maskgit, diffusion,applestable,qiu2025refining, yu2026enhancinggeometricperceptionvlms}. LVM \citep{LVM}, an auto-regressive generative model, effectively converted visual information into language-like visual sentences and improved understanding capabilities. MAE and Point-MAE \citep{he2022masked,point_mae}, utilizing a reconstruction strategy, established unified visual architectures across various downstream tasks in 2D and 3D domains. The ability for ICL has also been demonstrated within foundation models, as researchers employed specialized training methodologies \citep{bar2022visual, fang2024explore, wang2023seggpt, wei2025learning,yue2024addme, yu2025omnialpha0, yue2023chatface} to endow these models with superior in-context learning capabilities, thus providing a robust foundation for the domain of Visual ICL (VICL).

\subsection{Visual In-Context Learning via In-Painting} \label{subsec:vicl}

MAE-VQGAN \citep{bar2022visual} and Painter \citep{wang2023images} serves as crucial in-painting backbones for VICL, with copious works building upon and enhancing this framework.
Existing work \citep{zhang2024makes,sun2023exploring,xu2024towards} primarily focuses on retrieval to obtain better prompt. 
\citet{sun2023exploring} studied prompt spatial arrangement, testing eight configurations and reporting improved results through voting.
\citet{zhang2024instruct} pioneered visual prompt tuning \citep{adapter,hu2021lora,liu2023pre,bahng2022exploring}, adding a noise border to prompts. 
\revise{PANICL \citep{zhang2025panicl} employs a training-free k-nearest-neighbor fusion integrates multiple prompts to alleviate the bias inherent in single prompt.
PICO \citep{jiang2025personalized} reformulates personalized vision problem under the VICL paradigm and exhibits clear advantages.}
\citet{taskvector} focused on identifying task vector activates backbone to optimize VICL process.
\condenser{} \citep{condenser} leveraged prompt composition \cite{tsca} to aggregate informative cues from multiple prompts.
However, the patch-wise information aggregation strategy in \condenser{} \citep{condenser} exhibits inherent limitations, and its supervision over exemplar quality remains insufficiently comprehensive.
Motivated by these, we propose a locality fusion scheme coupled with three cooperative objectives, establishing a more reliable paradigm for prompt fusion.

\section{Method: \modelname{}}
\label{sec:method}

\subsection{Problem Formulation and Method Overview}
\label{subsec:overview}
Given a prompt database, $\mathcal{D} = \{P_i \}_{i=1}^{|\mathcal{D}|}$, where each prompt comprises an image-label pair. 
The pixel-level retriever $\mathcal{R}$ identifies top-$\Number$ similar prompt pairs $ \mathcal{P}=\{P_{_\numindex} = (X_n,Y_n)\}_{\numindex=1}^\Number$, for a given query image $X_q \in \mathbb{R}^{H \times W \times 3}$. 
Following previous settings, we adopt MAE-VQGAN, configured with patch size of 16 and feature dimension of $D$, as the backbone. 
Under general setting, the closest pair $P_1=(X_1,Y_1)\in \mathbb{R}^{H\times2W\times3}$ provides the prompt. We concatenate prompt $P_1$ with the query image $X_q$ to construct the canvas
$ S_1 = \begin{bmatrix}
  X_1 & Y_1 \\
  X_q & [M] \\
\end{bmatrix} $, where $[M]$ denotes the mask need to be recovered. The backbone output at the masked location corresponds to the $X_q$'s predicted label $\hat{Y_q}$.

\begin{figure*}[t]
    \centering
    \includegraphics[width=\textwidth]{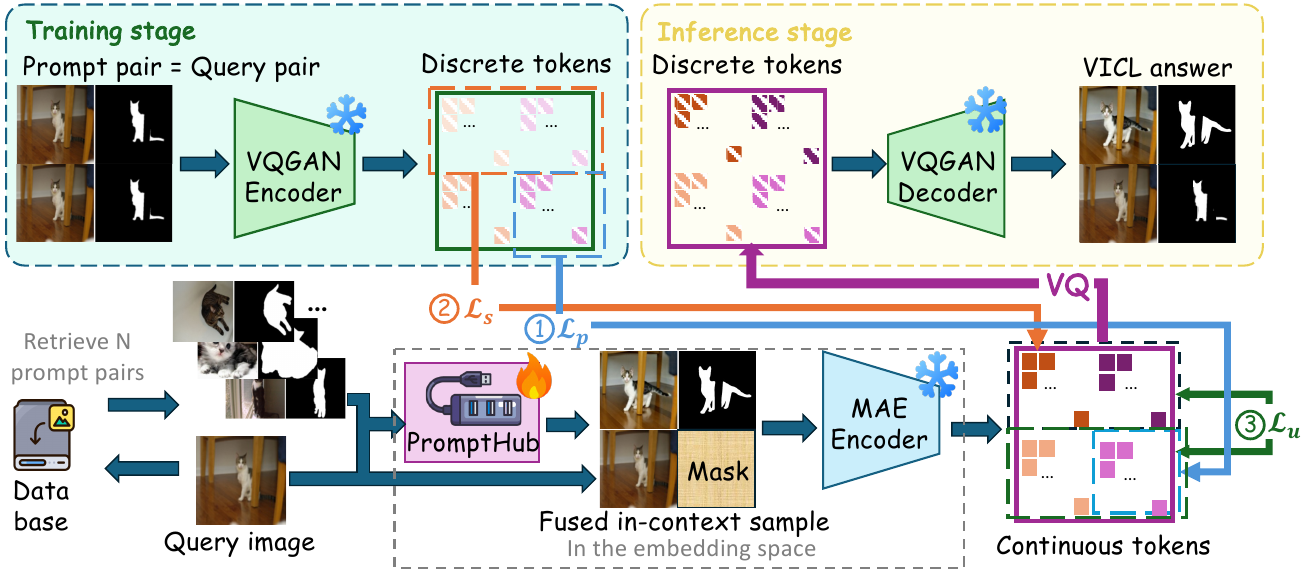}
    \caption{The training and inference framework of \modelname{} based on MAE-VQGAN. 
    }
    \label{fig:arc}
\end{figure*}

In our framework, $N$ prompt pairs $\mathcal{P}$ and query image $X_q$ are processed by the query-adaptive \modelname{} module, producing fused features $F_{X_f},F_{Y_f} \in \mathbb{R}^{\frac{H}{16} \times \frac{W}{16}\times D}$. $F_{X_f},F_{Y_f}$ and query image features $F_{X_q}$, mask features $F_{[M]}$ are concatenated into the canvas $S_f = \begin{bmatrix}
  F_{X_f} & F_{Y_f} \\
  F_{X_q} & F_{[M]} \\
\end{bmatrix} $. Then $S_f$ is passed through MAE-VQGAN, excluding the patch embedding, to generate the VICL answer.

\subsection{\modelname{} Module Design}

\begin{figure}[t]
\centering
\begin{minipage}[t]{0.6\textwidth}
    \centering

    \includegraphics[width=\textwidth]{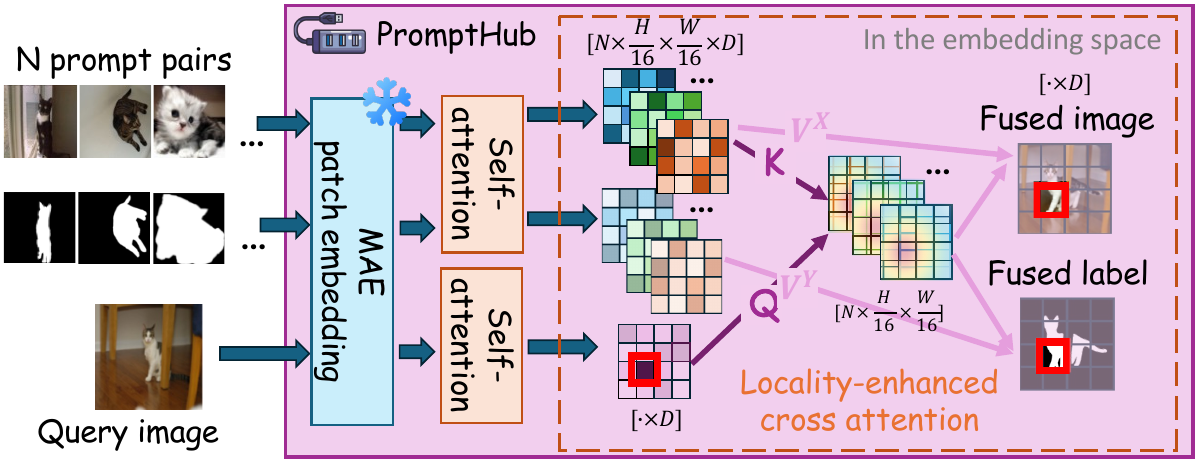}
    \vspace{-1.5em}
    \caption{
    \textbf{\modelname{} module design.} 
    \(N\) prompt pairs and query image are embedded into the MAE patch space, where locality-enhanced fusion integrates spatially cues into a fused prompt aligned with query’s informative content.
    }
    \label{fig:prompthub}
\end{minipage}%
\hfill
\begin{minipage}[t]{0.36\textwidth}
    \centering
    \includegraphics[width=\textwidth]{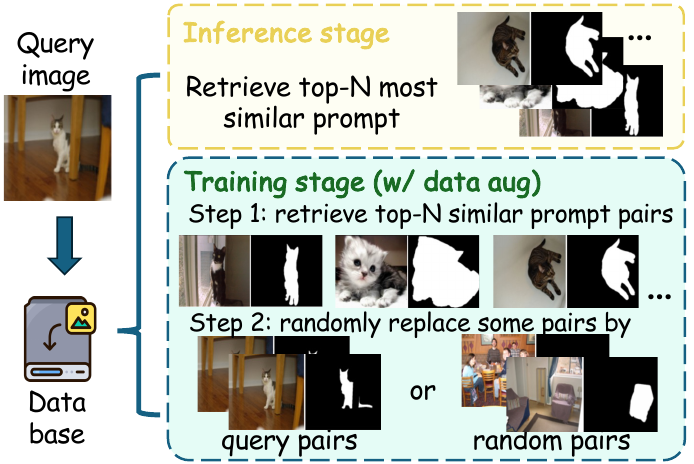}
    \vspace{-1.5em}
    \caption{
    \textbf{Data augment of \modelname{}}. 
    In training, the top-$N$ pairs are randomly substituted with either query pairs or random pairs.
    }
    \label{fig:data_aug}
\end{minipage}
\end{figure}

To expand the receptive field during fusion while mitigating the impact of boundary noise, we employ a locality-enhanced prompt fusion strategy.
\modelname{} locally fuse $\mathcal{P}$ into a unified prompt pair $(F_{X_f}, F_{Y_f})$ in the embedding space. 
The workflow is shown in \Cref{fig:prompthub}.

We first process images $(X_q,\mathcal{P})$ by embedding layer to yield $E_{X_q}, E_{X_{1:N}}, E_{Y_{1:N}} \in \mathbb{R}^{\frac{H}{16} \times \frac{W}{16} \times D}$.

Subsequently, we deploy a self-attention transformation $\mathrm{SA}(\cdot)$ to align the query and prompts to similar patterns, thereby generating the resultant features $F_{X_q}, F_{X_{1:N}}, F_{Y_{1:N}}$.

Thereafter, we use a query-adaptive locality-enhanced cross-attention to extract spatial information from prompt pairs features $(F_{X_{1:N}}, F_{Y_{1:N}})$, which achieves the fused exemplar features $(F_{X_f}, F_{Y_f})$. 

We define the locality prior as a probability distribution controlled by the hyper-parameter $\sigma$. To effectively represent this locality distribution, we employ either Gaussian prior or Laplacian prior, with the choice governed by a hyperparameter:
\begin{gather}
\psi(h,w,x,y) = 
\begin{cases} 
\exp\left(-\frac{(x-h)^2 + (y-w)^2}{2\sigma^2}\right), & \text{Locality prior} =\text{Gaussian prior} \\
\exp\left(-\frac{\sqrt{(x-h)^2 + (y-w)^2}}{\sigma}\right), & \text{Locality prior} = \text{Laplacian prior}
\end{cases}.
\end{gather}

For each query image token $F_{X_q}[h,w]$, it has a specific locality matrix ${\Psi}_{h,w}$ centered at $(h,w)$:
\begin{equation}
{\Psi}_{h,w} = 
\begin{bmatrix}
\psi(h, w, 1, 1) & \cdots & \psi(h, w, 1, \frac{W}{16}) \\
\vdots  & \ddots & \vdots \\
\psi(h, w, \frac{H}{16}, 1) & \cdots & \psi(h, w, \frac{H}{16}, \frac{W}{16})
\end{bmatrix}.
\end{equation}

During the VICL inference phase, no matched query label $Y_q$ is available for constructing fused prompt label $F_{Y_f}$. However, the specific correspondence still exists between prompt images $X_{1:N}$ and prompt labels $Y_{1:N}$. Therefore, we share the prompt images features $F_{X_{1:N}}$ as the key in the attention mechanism, compute the generalized attention scores, and subsequently perform localized weighting to procure locality-enhanced attention weights $A_{h,w} \in \mathbb{R}^{N\times\frac{H}{16}\times \frac{W}{16}}$ for $F_{X_f}[h,w]$:
\begin{gather}
A_{h,w} =\mathrm{softmax}\left( \frac{\left(F_{X_q}[h, w] \times W_Q\right)\times \left(F_{X_{1:N}} \times W_K\right)^\top}{\sqrt{D}} \cdot {\Psi}_{h,w} \right),
\end{gather}

which $\cdot$ denotes element-wise multiplication, $\times$ denotes matrix multiplication, and $W_Q, W_K \in \mathbb{R}^{D \times D}$ represent the projection layers for mapping $Q$ and $K$ in the attention mechanism.

Ultimately, we multiply the locality-enhanced attention weights with the features $F_{X_{1:N}}$ and $F_{Y_{1:N}}$ through linearly transformed  to obtain the fused prompt pair features $F_{X_f}, F_{Y_f} \in \mathbb{R}^{\frac{H}{16} \times \frac{W}{16} \times D}$. $W_{V^X}, W_{V^Y} \in \mathbb{R}^{D \times D}$ denote linear layers in attention mechanism for image and label, respectively.
\begin{gather}
F_{X_f}[h, w] = A_{h,w} \times \left(F_{X_{1:N}} \times W_{V^X}\right) ,\quad
F_{Y_f}[h, w] = A_{h,w} \times \left(F_{Y_{1:N}} \times W_{V^Y}\right) .
\end{gather}

\subsection{Learning Objectives}
We introduce three complementary learning objectives to guide fusion module's training, collectively strengthening \emph{``fusion–utilization–prediction"} closed-loop for robust VICL, as illustrated in \Cref{fig:arc}.

\textbf{(i) Ensuring label prediction performance.} \quad Following \condenser{} and InMeMo, we also adopt a label prediction loss as the fundamental objective to preserve VICL’s contextual prediction behavior. Without this base supervision, parameterized VICL paradigms cannot function properly. Upon deriving the fused in-context sample $S_f$, we propagate it through the MAE encoder, generating a canvas of continuous tokens $\begin{bmatrix}
  T_{X_f}^c & T_{Y_f}^c \\
  T_{X_q}^c & T_{[M]}^c \\
\end{bmatrix} \in \mathbb{R}^{\frac{2H}{16}\times \frac{2W}{16} \times D}$ . These tokens are calibrated during pretraining to correspond with the VQGAN codebook space. Simultaneously, We construct the target canvas $S_q=\begin{bmatrix}
    X_q &Y_q\\
    X_q &Y_q\\
\end{bmatrix}\in \mathbb{R}^{2H \times 2W \times 3}$ by integrating the query pair as a prompt pair and process it through the VQGAN encoder, obtaining the corresponding discrete tokens $\begin{bmatrix}
    T_{X_q}^{d(1)} & T_{Y_q}^{d(1)}\\
    T_{X_q}^{d(2)} & T_{Y_q}^{d(2)}
\end{bmatrix} \in {\{1,2,...,N_c\}}^{\frac{2H}{16}\times \frac{2W}{16}}$ from the codebook.
Here, $N_c$ denotes the size of the codebook space, with $N_c = D$. $T_{X_q}^{d(1)}$ and $T_{Y_q}^{d(1)}$ represent the discrete tokens output as prompt, while $T_{X_q}^{d(2)}$ and $T_{Y_q}^{d(2)}$ correspond to the discrete tokens output as query. To optimize the label prediction results, we align the bottom-right portion $T_{[M]}^{c}$, which will be reconstructed by the VQGAN decoder, with the target $T_{Y_q}^{d(2)}$ using a cross-entropy loss. Here, $\mathcal{L}_{p}$ denotes the loss function for \emph{label prediction}:
\vspace{-0.25em}
\begin{equation}
    \mathcal{L}_p = -\mathbb{E}_{(h, w) \sim \mathcal{U}([1, \frac{H}{16}] \times [1, \frac{W}{16}])} \log T_{[M]}^{c}\left[h,w,T_{Y_q}^{d(2)}\right].
\end{equation}

\textbf{(ii) Fused-prompt feature alignment.} \quad 
The backbone tends to produce accurate predictions when exposed to the same prompt as the query.
We employ a cross-entropy alignment between the continuous tokens derived from the fused prompt pair $(T_{X_f}^c ,T_{Y_f}^c)$ and the discrete tokens corresponding to the query pair as a prompt $(T_{X_q}^{d(1)},T_{Y_q}^{d(1)})$ to make fused prompt pair closely approximate the query pair. The semantic integrity loss for improved \emph{fusion} is denoted as $\mathcal{L}_s$.
\vspace{-0.25em}
\begin{equation}
    \mathcal{L}_s = -\mathbb{E}_{(h, w) \sim \mathcal{U}([1, \frac{H}{16}] \times [1, \frac{W}{16}])} \left( \log T_{X_f}^{c}[h,w,T_{X_q}^{d(1)}] +\log T_{Y_f}^{c}[h,w,T_{Y_q}^{d(1)}] \right).
\end{equation}

\textbf{(iii) Enhance fused prompt utilization.} \quad 
Owing to discrepancy between fused prompt and query pair, backbone may regard useful prompt as unreliable and instead rely on its own capacity. 
We employ a cosine-similarity loss $\mathcal{L}_u$, designed to reduce the dissimilarity between query pair $(T_{X_q}^c,T_{[M]}^c)$ and fused prompt $(T_{X_f}^c,T_{Y_f}^c)$, thereby enhancing the backbone’s \emph{utilization} of fused prompt.
\vspace{-0.25em}
\begin{equation}
    \mathcal{L}_u = -\mathbb{E}_{(h, w) \sim \mathcal{U}([1, \frac{H}{16}] \times [1, \frac{W}{16}])} \left( \cos \left( T_{X_f}^{c}\left[h,w\right],T_{X_q}^{c}\left[h,w\right] \right)+\cos \left( T_{Y_f}^{c}[h,w],T_{[M]}^{c}[h,w]\right) \right).
\end{equation}

\quad We adopt $\lambda$ and $\gamma$ to balance diffenent losses. Let $\theta$ denote the parameters of \modelname{}. The ultimate synergistic optimization objective is formulated as:
\vspace{-0.25em}
\begin{equation}
    \min_{\theta} \; \mathcal{L}_p + \lambda \mathcal{L}_s + \gamma \mathcal{L}_u.
\end{equation}
\vspace{-1em}

\subsection{Retrieve Scheme for Data Augment}

In the inference phase, we consistently retrieve the top-$N$ most similar prompt pairs $\mathcal{P} = \{P_n\}_{n=1}^N$ from the database $\mathcal{D}$, utilizing the most relevant raw prompt pairs for improved VICL. 

During training, we employ a data augmentation strategy to enhance two regularization objectives' effect. 
Based on the retrieved top-$N$ prompt pairs $\mathcal{P} = \{P_n\}_{n=1}^N$, we might replace some prompt pairs $P_n$ with either query pairs $P_q=(X_q,Y_q)$ or randomly retrieved pairs $P_r$, as shown in \Cref{fig:data_aug}.

\textbf{(i)} Substitute with query pair to better utilize the fused prompt. \quad
Under typical settings, defining prompt pair $P_1$ as query pair $P_q$ generally yields minimal discrepancy. To this end, we replace current prompt pairs $P_n$ with query pairs $P_q$ with probability $p_q$. This substitution establishes a purified learning objective that minimizes discrepancy as much as possible, hence enhancing $\mathcal{L}_u$.

\textbf{(ii)} Substitute with random pair to enhance \modelname{}'s robustness.  With probability $p_r$, we substitute prompt pair $P_n$ with a randomly retrieved pair $P_r$, introducing a controlled level of noise that enhances $\mathcal{L}_s$ and \modelname{}'s stability. This technique guarantees when high-quality prompts are unavailable during inference, \modelname{} retains its capacity to achieve robust VICL results.

\section{Experiments}
\label{sec:experiments}

\subsection{Experimental Setup}
\label{subsec:setup}

\textbf{Downstream Tasks and Datasets.} \quad To ensure a fair comparison, we employ three well-established tasks foreground segmentation, single-object detection, and colorization along with their associated datasets, within the domain of VICL. For \textbf{foreground segmentation}, we employ Pascal-5$^i$ \citep{shaban2017one}, which consists of four folds, with each fold containing data from five different classes. We conduct experiments across all folds and analyze the results by presenting the mean intersection over union (mIoU) for each fold. In the case of \textbf{single-object detection}, we utilize the Pascal VOC2012 \citep{everingham2015pascal} dataset, also employing mIoU as the evaluation metric. For the \textbf{coloring task}, we randomly select 50,000 images from the ImageNet-1K ILSVRC2012 \citep{russakovsky2015imagenet} training set, with 50 images chosen from each of the 1,000 classes to form the label portion of our training set. The 50,000 images from the validation set of ImageNet-1K ILSVRC2012 are used as the label portion of our test set. We convert training set and test set label portion to grayscale images, which served as the input queries. We use MSE as the evaluation metric.

\noindent\textbf{Implementation Details.} \quad We adopt MAE-VQGAN \citep{bar2022visual} as the backbone architecture and utilize Prompt-SelF's \citep{sun2023exploring} pixel-level retriever for prompt retrieval. During training, we use the training set as the database for prompt pairs while also employing the training set as the query. 
In the testing phase, the validation set serves as the query collection, while the training set acts as the database. The input image resolution to the model is \(224 \times 224\), with each sub-image having a resolution of \(112 \times 112\). We utilized Gaussian prior as the default locality prior.

\noindent\textbf{Training Configurations.} \quad We employed SGD optimizer with a learning rate initialized at 0.04, which decays according to cosine annealing warm restarts scheduler.
For segmentation and detection tasks, training is performed for 100 epochs, while coloring task requires 10 epochs. The corresponding $\sigma$ values for foreground segmentation, object detection, and colorization tasks are 0.65, 0.5, and 2.5, respectively. 
Hyper-parameter $\lambda$ is set to 0.5, and $\gamma$ is set to 0.2. The experiments were performed on single 80G A100 GPUs with a batch size of 16.

\begin{table}[t]
    \centering
    \vspace{-2em}
    \caption{
    \modelname{} performance is compared with different baselines in three downstream tasks foreground segmentation (\textbf{Seg}.), single-object detection (\textbf{Det}.), and image colorization (\textbf{Col}.). The results for $\Number = 1,16$, representing the cases with 1 and 16 prompts respectively, are listed separately. The highest results are denoted in \textbf{bold}, while the suboptimal results are indicated in \textit{italics}. }
    \resizebox{\columnwidth}{!}{
        \begin{tabular}{lccccccc}
        \toprule
        & \multicolumn{5}{c}{\textbf{Seg. (mIoU ↑)}} &  &   \\
        \multirow{-2}{*}{\textbf{Model}} & \textbf{Fold-0} & \textbf{Fold-1} & \textbf{Fold-2} & \textbf{Fold-3} & \textbf{Mean} & \multirow{-2}{*}{\textbf{Det. (mIoU ↑)}} & \multirow{-2}{*}{\textbf{Col. (MSE ↓)}} \\
        \midrule
        \midrule
        \rowcolor[HTML]{F2F2F2} \textbf{\emph{Zero-Shot}} & & & & & & & \\
        Random \citep{bar2022visual} & 28.66 & 30.21 & 27.81 & 23.55 & 27.56 & 25.45 & 0.67 \\
        UnsupPR \citep{zhang2024makes} & 34.75 & 35.92 & 32.41 & 31.16 & 33.56 & 26.84 & 0.63 \\
        Prompt-SelF \citep{sun2023exploring} & 35.69 & 38.25 & 35.86 & 33.37 & 35.79 & 28.08 & 0.63 \\
        \midrule
        
        \rowcolor[HTML]{F2F2F2} \textbf{\emph{Retriever Training}} & & & & & & & \\
        SupPR \citep{zhang2024makes} & 37.08 & 38.43 & 34.40 & 32.32 & 35.56 & 28.22 & 0.63 \\
        Partial2Global \citep{xu2024towards} & 38.81 & 41.54 & 37.25 & 36.01 & 38.40 & 30.66 & 0.58 \\
        \midrule
        \rowcolor[HTML]{F2F2F2} \textbf{\emph{PEFT}} & & & & & & & \\
        InMeMo \citep{zhang2024instruct} & 41.65 & 47.68 & 42.43 & 40.80 & 43.14 & 43.21 & - \\
        \midrule
        \rowcolor[HTML]{F2F2F2} \textbf{\emph{Task Vectors}} & & & & & & & \\
        VTV \citep{taskvector} & 38.00 & 37.00 & 33.00 & 32.00 & 33.50 & - & - \\
        \midrule
        \rowcolor[HTML]{F2F2F2} \textbf{\emph{Prompt Fusion}} & & & & & & & \\
        \rowcolor[HTML]{F0FFFF} \condenser{}$_{\Number=1}$ \citep{condenser}& 42.13 & 50.31 & 42.20 & 41.90 & 44.14 & 43.22 & 0.560 \\
        \rowcolor[HTML]{ADD8E6} \condenser{}$_{\Number=16}$ \citep{condenser}& \emph{45.53} & \emph{52.06} & \emph{44.33} & \emph{44.58} & \emph{46.63} & \emph{44.64} & 0.539 \\
        \midrule
        \rowcolor[HTML]{F0FFFF} \modelname{}$_{\Number=1}$ \textbf{\emph{(Ours)}} & 43.26 & 50.75 & 43.83 & 42.82 & 45.17 & 44.51 & \emph{0.533} \\
        \rowcolor[HTML]{ADD8E6} \modelname{}$_{\Number=16}$ \textbf{\emph{(Ours)}}& \textbf{45.93} & \textbf{53.12} & \textbf{45.44} & \textbf{46.74} & \textbf{47.81} & \textbf{45.59} & \textbf{0.503} \\

        \hline
        \bottomrule
        \end{tabular}
    }
    \vspace{-1.5em}
    \label{tab:main_table}
\end{table}

\subsection{Comparison with State-of-The-Arts}
\label{subsec:sota}

\noindent\textbf{Baselines.} \quad We compare our method against comprehensive state-of-the-art approaches built on the MAE-VQGAN framework. Our competitors are categorized into four groups: (1) Zero-shot methods, including MAE-VQGAN \citep{bar2022visual} and UnsupPR \citep{zhang2024makes} and Prompt-SelF \citep{sun2023exploring}, which do not require additional retriever training; (2) Methods that necessitate retriever training, such as SupPR \citep{zhang2024makes} and Partial2Global \citep{xu2024towards}; 
(3) Approach that leverages prompt tuning, exemplified by InMeMo \citep{zhang2024instruct}; 
(4) Method of finding and utilizing the task vector VTV \citep{taskvector}.
(5) Method that employs prompt fusion, \condenser{} \citep{condenser}, to enable multi-prompt VICL, with comparisons reported under both single-prompting and multi-prompting settings.

\noindent\textbf{(i) Performance on Standard Tasks.} \quad 
\Cref{tab:main_table} demonstrates that \modelname{} achieves consistent improvements across all tasks under both single-prompt and multi-prompt settings. In single-prompt scenario, \modelname{} surpasses \condenser{} by 2.3\%, 3.0\%, and 5.1\% on segmentation, detection, and colorization, respectively. Under multi-prompt scenario, it further attains gains of 2.5\%, 2.1\%, and 7.2\% on same tasks.
\modelname{}'s output visualization is discussed further in the appendix.

\begin{wraptable}{r}{0.45\textwidth}
    \centering
    \vspace{-2em}
    \caption{Transferability evaluation. We train models on COCO-5$^i$ and test on Pascal-5$^i$.}
    \resizebox{0.45\textwidth}{!}{
    \begin{tabular}{l c c c c c}
        \toprule
        \multirow{2}{*}{\textbf{Model}} & \multicolumn{5}{c}{\textbf{Seg. (mIoU ↑)}} \\
        & \textbf{Fold-0} & \textbf{Fold-1} & \textbf{Fold-2} & \textbf{Fold-3} & \textbf{Mean} \\
        \midrule
        \midrule
        Prompt-SelF & 40.13 & 42.14 & 37.84 & \textbf{38.52} & 39.66  \\
        InMeMo & 38.74 & 43.82 & 40.45 & 37.12 & 40.03  \\
        \condenser{}$_{\Number=1}$ & \emph{40.39} & 44.54 & 40.23 & 36.33 & 40.37 \\
        \condenser{}$_{\Number=16}$ & 40.37 & \emph{44.85} & 
        \emph{41.03} & 35.84 & 40.52 \\        
        \rowcolor[HTML]{F0FFFF} \modelname{}$_{\Number=1}$ & \text{40.36} & 45.24 & 40.43 & \emph{37.94} & \emph{41.00}  \\
        \rowcolor[HTML]{ADD8E6} \modelname{}$_{\Number=16}$ & \textbf{42.69} & \textbf{46.71} & \textbf{41.97} & \text{37.31} & \textbf{42.17}  \\
        \bottomrule
    \end{tabular}
    }
    \vspace{-1em}
    \label{tab:ada}
\end{wraptable}

\noindent\textbf{(ii) Performance on Domain Adaption Task.} \quad
In real-world applications, the data for inference 
may undergo domain adaptation compared to the training data. Thus, testing the transferability of different VICL schemes is crucial. We trained the PromptHub on the COCO-$5^i$ \citep{lin2014microsoft} using the same settings as previous works \citep{condenser,sun2023exploring,zhang2024instruct}, 
and evaluate it on the Pascal-$5^i$. 
As shown in \Cref{tab:ada}, \modelname{} 
\begin{wrapfigure}[]{r}{0.5\textwidth}
    \centering
    \vspace{-1em}    \includegraphics[width=0.5\columnwidth]{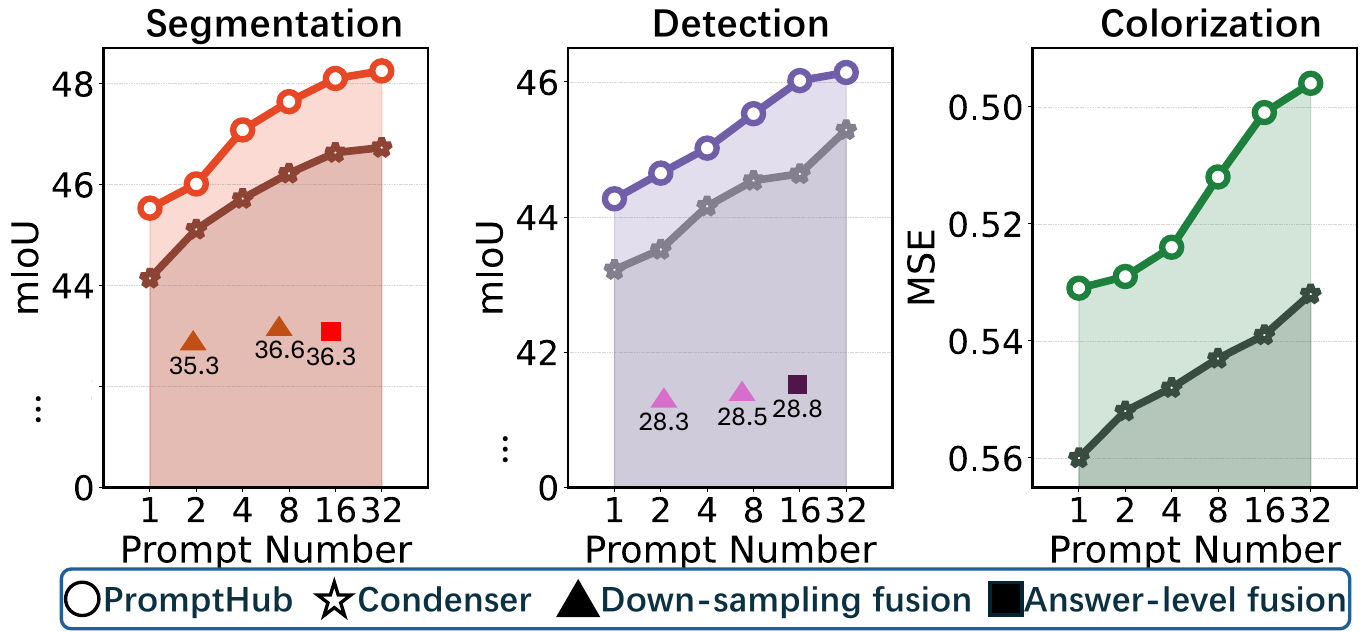}
    \vspace{-2em}
    \caption{
    Performance comparison with baselines in multi-prompt VICL scenario.
    }
    \vspace{-2em}
\label{fig:K_influence}
\end{wrapfigure}

demonstrates substantially larger improvements than other baselines, outperforming \condenser{} by 4.1\% in the multi-prompt setting, highlighting the strong transferability.

\textbf{(iii) Performance under the multi-prompting scenario.} \quad 
To validate the scalability of PromptHub, we compare it with \condenser{} under various $\Number$, specifically 1, 2, 4, 8, 16, and 32. In addition, we report results under down-sampling$_{N=2,N=7}$ \citep{zhang2024makes} and answer-level$_{N=16}$ \citep{sun2023exploring}. The experimental results demonstrate our approach not only improves performance as $N$ increases, but also consistently surpasses other baselines by a large margin, as shown in \Cref{fig:K_influence}.

\begin{table}[t]
\centering

\caption{Ablation study of \modelname{}. The best are marked in \textbf{bold} and second-best in \emph{italic}.}

\resizebox{0.92\columnwidth}{!}{
\begin{tabular}{llcccccc}
\toprule
& & \multicolumn{5}{c}{\textbf{Seg. (mIoU ↑)}} &\\
\multirow{-2}{*}{\textbf{\#}} &
  \multirow{-2}{*}{\textbf{Model}}  &
  \textbf{Fold-0} &
  \textbf{Fold-1} &
  \textbf{Fold-2} &
  \textbf{Fold-3} &
  \textbf{Mean} &
  \multirow{-2}{*}{\textbf{Det. (mIoU ↑)}} \\
\hline
\midrule
\rowcolor[HTML]{F0FFFF} (0)  & \modelname{}$_{\Number=1}$ & 43.26 & 50.75 & 43.83 & 42.82 & 45.17 & 44.51 \\
\rowcolor[HTML]{ADD8E6} (1)  & \modelname{}$_{\Number=16}$ & \emph{45.93} & \textbf{53.12} & 45.44 & \textbf{46.74} & \textbf{47.81} & \textbf{45.59} \\
\midrule
\rowcolor[HTML]{F2F2F2} \multicolumn{6}{l}{\textbf{\emph{Effectiveness of Learning Objectives}}} & &\\
(2)  & w/o $\mathcal{L}_{u}$ $_{N=1}$         & 42.71 & 51.14  & 42.78 & 42.41 & 44.76 & 43.45 \\
(3)  & w/o $\mathcal{L}_{u}$ $_{N=16}$ &     45.54 & 52.25 & 44.59 & 44.47 & 46.71  & 44.83 \\
(4)  & w/o $\mathcal{L}_{s}$ $_{N=1}$         & 42.23 & 50.52  & 42.29 & 42.16 & 44.30 & 43.12 \\
(5)  & w/o $\mathcal{L}_{s}$ $_{N=16}$    &    44.72 & 51.77 & 43.57 & 43.30 & 45.84  & 44.27 \\
(6)  & w/o $\mathcal{L}_{p}$ $_{N=1}$      & 8.51 & 10.13  & 9.46 & 8.33 & 9.11 & 13.23 \\
(7)  & w/o $\mathcal{L}_{p}$ $_{N=16}$ &     9.41 & 13.44 & 12.29 & 10.62 & 11.44  & 12.87 \\
\midrule
\rowcolor[HTML]{F2F2F2} \multicolumn{6}{l}{ \textbf{\emph{Effectiveness of Locality-Enhanced Fusion}}}  & & \\
(8)  & w/ Laplacian Prior$_{\Number=1}$& 43.74& 50.93& 43.51& 43.05 & 45.31 & 43.93\\
(9)  & w/ Laplacian Prior$_{\Number=16}$ & \textbf{46.42} & \emph{52.87} & \textbf{45.45} & \emph{46.16} & \emph{47.72} & \emph{45.47}\\
(10)  & Global Fusion$_{\Number=1}$& 41.77& 49.04& 42.69& 40.73& 43.55 & 41.86\\
(11)  & Global Fusion$_{\Number=16}$ & 41.91& 50.45& 43.76& 42.43 &  44.64 &42.49\\
(12)  & Convolution-Based Fusion$_{\Number=1}$   & 42.56 & 50.15 & 42.79  & 42.52 & 44.51   & 43.83 \\
(13)  & Convolution-Based Fusion$_{\Number=16}$     & 45.28 & 51.68 & 45.34  & 45.51 & 46.95   & 45.07 \\
\midrule
\rowcolor[HTML]{F2F2F2} \multicolumn{6}{l}{ \textbf{\emph{Effectiveness of Data Augment Technique}}}  & &\\
(14)  & w/o Data Augment$_{\Number=1}$ & 43.11 & 51.22 & 43.17 & 42.34 & 44.96  & 43.52 \\
(15)  & w/o Data Augment$_{\Number=16}$  & 45.84 & 52.01 & 44.83 & 45.60 & 47.07 & 45.06 \\
\hline

\bottomrule
\end{tabular}
}
\vspace{-1em}
\label{tab:abla}
\end{table}

\begin{figure*}[t]
    \centering
    \includegraphics[width=0.92\textwidth]{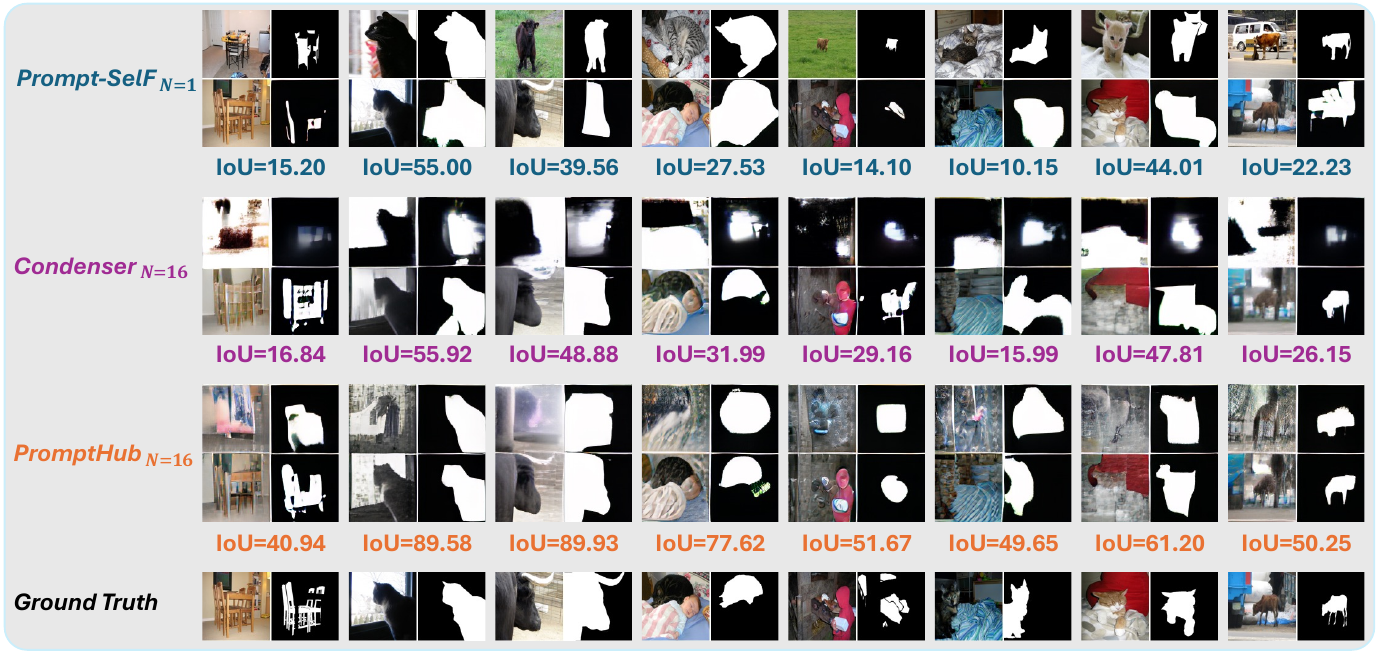}
    \vspace{-1em}
    \caption{The visualization of the fused prompt pair after passing  through the VQGAN decoder.
    \vspace{-10em}
    }
\label{fig:model_learns_img}
\end{figure*}
\vspace{-0.5em}

\subsection{Model Analysis}
\label{subsec:model_analyses}

For a comprehensive ablation study, we designed several variants, as summarized in \Cref{tab:abla}, where Variants (0) – (1) correspond to the canonical configurations.

\noindent \textbf{(i) Effectiveness of Learning Objectives.} \quad To comprehensively evaluate the contributions of each learning objective, we conducted an ablation analysis by individually removing the three objectives. 
The experimental results demonstrate \emph{``fusion-utilization-prediction"} objectives are mutually complementary, and omitting any of them leads to performance degradation in multi-prompt VICL. In particular, the primary objective, label prediction $\mathcal{L}_p$, is indispensable for preserving VICL’s contextual prediction behavior; \emph{without it, the training-based VICL paradigm with additional parameters cannot function effectively}. Meanwhile, $\mathcal{L}_s$ and $\mathcal{L}_u$ act as crucial regularization terms, ensuring fused exemplars' quality and the backbone’s effective utilization. \emph{The absence of either damages the pipeline in VICL and results in mediocre performance.}

\noindent \textbf{(ii) Effectiveness of Locality-Enhanced Prompt Fusion.} \quad 
We compare locality-enhanced fusion with global fusion, patch-wise fusion (\condenser{} \citep{condenser}), and convolution-based fusion, where the latter replaces the spatial prior with convolutional transformations.
\emph{Notably, locality-enhanced fusion can be viewed as a higher-level framework, within which global fusion and patch-wise fusion emerge as two complementary instantiations, corresponding to larger and smaller values of the locality parameter $\sigma$, respectively.}
As shown in \Cref{tab:abla}, both types of locality priors achieve superior performance.
\emph{Upon observation, maintaining an appropriate balance between global receptive fields and spatial locality proves essential. The locality accords with the fusion principle that enriches information capture while mitigating long-range noise.}

\textbf{(iii) Effectiveness of Data Augment Technique.} \quad
We conducted experiments under scenarios without data augmentation, only utilizing the top-$N$ prompt pairs for fusion during training, as illustrated in Variants (14) – (15). 
The results indicate that removing data augmentation diminishes the performance of \modelname{} in VICL tasks, confirming the effectiveness of data augmentation. It better reinforce fused prompt utilization and enhances noise resistance.

\begin{wrapfigure}[]{r}{0.45\textwidth}
\centering
    \vspace{-1em}
    \includegraphics[width=0.45\columnwidth]{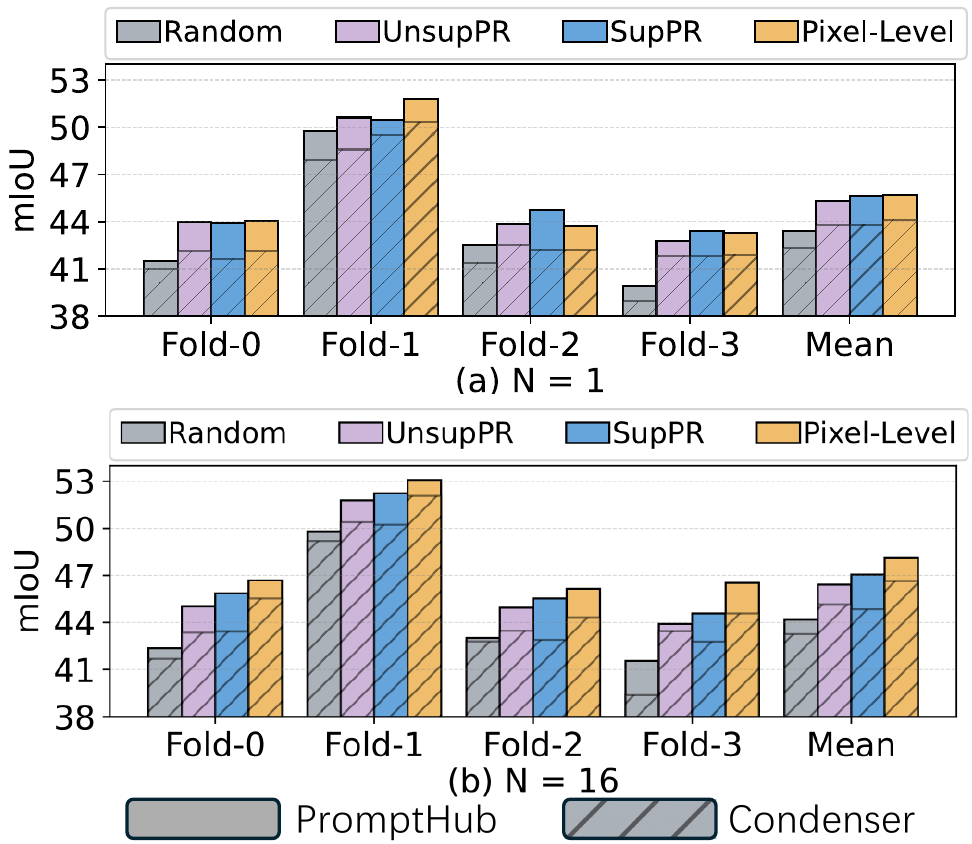}
    \vspace{-2em}
    \caption{Comparison of \modelname{} and \condenser{} across different retrieval.
    }
    \vspace{-1em}
\label{fig:dif_retri}
\end{wrapfigure}

\textbf{(iv) Impact on Different Retrievers.} \quad \emph{Exploring better prompt retrieval and investigating multi-prompt fusion are two orthogonal research directions, while the fusion plugin can be adapted to different retrievers.} We investigated performance of \modelname{} using different 
retrievers, as presented in \Cref{fig:dif_retri}. We evaluated four types of retrievers: random selection \citep{bar2022visual}, UnsupPR \citep{zhang2024makes}, SupPR \citep{zhang2024makes} and Pixel-Level retriever \citep{sun2023exploring}. Experimental results demonstrate \modelname{} is more effective than \condenser{} across all retrieval schemes, further highlighting its generalizability. Additionally, the performance of our method is influenced by the choice of retriever; pixel-level retrievers consistently deliver best results, underscoring the alignment between pixel-level retrieval and locality-aware design philosophy.

\textbf{\revise{(v) Transferability on Unseen Tasks.}} \quad
\revise{We evaluate cross-task transferability by training all models solely on segmentation (Pascal-5, four folds) and directly testing them on detection (Pascal VOC 2012) without any fine-tuning. We compare \modelname{} with the \condenser{} baseline using their released checkpoints, and report results in \Cref{tab:unseen}.}
\begin{wraptable}{r}{0.55\textwidth}
    \centering
    \vspace{-1em}
    \caption{\revise{Transferability experiment (unseen task) where both \condenser{} and \modelname{} are trained on segmentation and evaluated on detection.}}
    \resizebox{0.55\textwidth}{!}{
    \begin{tabular}{lccccc}
    \toprule
    & \multicolumn{5}{c}{\textbf{Det. (mIoU ↑)}}   \\
    \multirow{-2}{*}{\textbf{Method}} & \textbf{Fold-0} & \textbf{Fold-1} & \textbf{Fold-2} & \textbf{Fold-3} & \textbf{Mean} \\
    \midrule
    \midrule
    Condenser$_{N=1}$  & 38.15 & 35.70 & 35.49 & 30.31 & 34.91 \\
    Condenser$_{N=16}$ & 41.25 & 36.66 & 37.86 & 39.02 & 38.70  \\
    \rowcolor[HTML]{F0FFFF} PromptHub$_{N=1}$  & 41.59 & 36.25 & 37.71 & 32.15 & 36.93 \\
    \rowcolor[HTML]{ADD8E6} PromptHub$_{N=16}$ & 43.40 & 38.55 & 39.52 & 40.66 & 40.53  \\
    \bottomrule
    \end{tabular}
    }
    \vspace{-1em}
    \label{tab:unseen}
\end{wraptable}
\revise{PromptHub consistently surpasses Condenser in this challenging unseen-task setting. With $N=16$, PromptHub achieves a +1.83\% mIoU gain, indicating that our locality-aware fusion captures more robust and transferable visual cues than the patch-wise fusion used in Condenser.} 
\revise{\emph{We note that although the overall performance is strong, the training process is not fully task-agnostic. Since the model is trained to reconstruct segmentation masks, a domain gap naturally emerges when transferring to bounding box detection, which leads to a certain degree of performance drop.}
}

\textbf{\revise{(vi) Performance Evaluation under Spatial Misalignment.}} \quad
\revise{Spatial misalignment between prompts and queries may negatively affect prompt fusion performance. To evaluate performance}
\begin{wraptable}{r}{0.55\textwidth}
    \centering
    \vspace{-1.5em}
    \caption{\revise{Comparison of standard and perturbed mIoU under spatial misalignment, along with the corresponding performance drops.}}
    \resizebox{0.55\textwidth}{!}{
    \begin{tabular}{lccc}
    \toprule
    \textbf{Method} & \textbf{Standard mIoU}
    & \textbf{Perturbed mIoU} & \textbf{Performance Drop}\\
    \midrule
    \midrule
    Condenser$_{N=1}$  & 44.14 & 42.36 & -1.78 \\
    Condenser$_{N=16}$ & 46.63 & 45.24 & -1.39 \\
    \rowcolor[HTML]{F0FFFF} PromptHub$_{N=1}$  & 45.17 & 44.01 & -1.16 \\
    \rowcolor[HTML]{ADD8E6} PromptHub$_{N=16}$ & 47.81 & 47.15 & -0.66 \\
    \bottomrule
    \end{tabular}
    }
    \vspace{-1em}
    \label{tab:flip}
\end{wraptable}
\revise{under position shifts, we conducted an experiment where query pairs were horizontally flipped and retrained to simulate severe spatial misalignment between the retrieved prompts and the query image. We compared the performance drop of \condenser{} and \modelname{} under the perturbed conditions in \Cref{tab:flip}. \modelname{} is substantially more robust to spatial misalignment than \condenser{}. Its locality-aware fusion mitigates the sensitivity to positional shifts that affects \condenser{}'s patch-wise fusion. In addition, increasing number of prompts $N$ further reduces misalignment effects by improving chance of encountering better-aligned prompt pairs.}

\subsection{Discussion: What does \modelname{} Learn?}
In \Cref{fig:model_learns_img}, we present visualizations of Prompt-SelF, as well as fusion samples reconstructed through the VQGAN decoder for \condenser{}$_{N=16}$ and \modelname{}$_{N=16}$. Given that this visualization relies on reconstructed outputs, some bias may be inevitably introduced. We observe that in Prompt-SelF, label prediction often tends to be highly similar to the retrieved prompt label, leading to poor performance when the retrieved label show little similarity to ground-truth answer.
\begin{wraptable}{r}{0.55\textwidth}
    \centering
    \vspace{-1em}
    \caption{\revise{Comparison the mIoU between the fused prompt labels and the query labels across methods to evaluate semantic alignment.}}
    \resizebox{0.55\textwidth}{!}{
    \begin{tabular}{lccccc}
    \toprule
    & \multicolumn{5}{c}{\textbf{Seg. (mIoU ↑)}}   \\
    \multirow{-2}{*}{\textbf{Method}} & \textbf{Fold-0} & \textbf{Fold-1} & \textbf{Fold-2} & \textbf{Fold-3} & \textbf{Mean} \\
    \midrule
    \midrule
    Condenser$_{N=1}$  & 18.93 & 29.73 & 24.26 & 27.94 & 25.22 \\
    Condenser$_{N=16}$ & 14.27 & 24.56 & 18.85 & 20.56 & 19.56  \\
    \rowcolor[HTML]{F0FFFF} PromptHub$_{N=1}$  & 21.25 & 37.61 & 35.01 & 29.79 & 30.92 \\
    \rowcolor[HTML]{ADD8E6} PromptHub$_{N=16}$ & 25.22 & 43.22 & 36.07 & 30.41 & 33.73  \\
    \bottomrule
    \end{tabular}
    }
    \vspace{-1em}
    \label{tab:iou_fused_prompt}
\end{wraptable}
The fusion results of \condenser{} appear as noisy black-and-white patterns, which may be attributed to its model-agnostic feature matching and patch-wise attention that fail to generate smooth representations, offering only heuristic contributions to performance.
\emph{In contrast, the fused prompts produced by \modelname{} exhibit significantly better visual quality, with fused prompts showing high similarity to the query pairs and smooth textures, thereby confirming the advantages of the locality-aware design and offering a more \revise{reliable} and trustworthy solution for prompt fusion in VICL.} 
\revise{Furthermore, we quantitatively compare the mIoU between fused prompt labels and query labels for \condenser{} and \modelname{}, as shown in \Cref{tab:iou_fused_prompt}. \modelname{}$_{N=16}$ achieves a 72\% higher similarity to the ground truth compared with \condenser{}$_{N=16}$, demonstrating \modelname{} produces higher-quality and more semantically coherent fused prompts.}
Although fused prompts may exhibit a gap from realistic images due to lack of fidelity constraints in decoding stage, our primary goal is to guide VICL inference rather than to generate photorealistic images.

\section{Conclusions}
\label{sec:conclusion}

In this work, we introduced \modelname{}, an interpretability paradigm that realizes the chain-wide enhancements \emph{``locality fusion–utilization–prediction"} for multi-prompt VICL. \modelname{} balances spatial locality with global receptive fields, supervises the quality of fused samples, and enhances the backbone's utilization on integrated prompts. Extensive experiments across diverse tasks demonstrate clear improvements over previous methods.
Furthermore, \modelname{}'s superior transferability, robustness and generalizability further highlight its potential for extensive implementation in diverse scenarios. We finally visualize the fused prompts, the results outperform the patch-wise scheme and provide stronger interpretability for prompt fusion methods.

\paragraph{Acknowledgments}
We sincerely thank the anonymous reviewers and chairs for their efforts and constructive suggestions, which have greatly helped us improve the manuscript. 
This work is supported in part by the National Key R\&D Program of China (2022YFB4701400/4701402), the National Natural Science Foundation of China under grants 624B2088, 62576122, 62301189, 62571298, and in part by the SSTIC Grants KJZD20230923115106012, KJZD20230923114916032, and GJHZ20240218113604008.

\bibliography{main}
\bibliographystyle{iclr2026_conference}

\clearpage
\appendix

\section{Scope of LLM Usage}

To remain compliant with responsible LLM usage protocols, we limited the scope of LLMs to improving readability and grammar. Every scientific contribution, including the conceptual development, experimental design, and analytical validation, was independently carried out and confirmed by the authors, and we retain complete responsibility.

\section{Future Works and Limitations}

\subsection{\revise{White-Box Dependency}}

\revise{Like \condenser{}, \modelname{} requires access to the backbone’s parameters and gradients to train the fusion module, even though the backbone itself is frozen. This design has been instrumental in ensuring the framework's success and robustness in its current applications. But this makes scaling to very large models or closed-source models challenging, as full gradients may be inaccessible or too costly. While this enables superior performance, extending prompt fusion to black-box or gradient-free settings is a key direction for future work.}

\subsection{Extending Applicability to Linguistic and Multi-Modal Domains}

\modelname{} is designed for VICL tasks with constrained inputs, utilizing positional correspondences between query and label image patches for locality-enhanced prompt fusion. Building on its success in the visual domain, future work will expand its scope to multi-modal scenarios by exploring generalized mechanisms that effectively align visual and linguistic modalities, enabling broader applicability and integration.

\begin{wrapfigure}[]{r}{0.5\textwidth}
\centering
    \vspace{-2em}
    \includegraphics[width=0.8\linewidth]{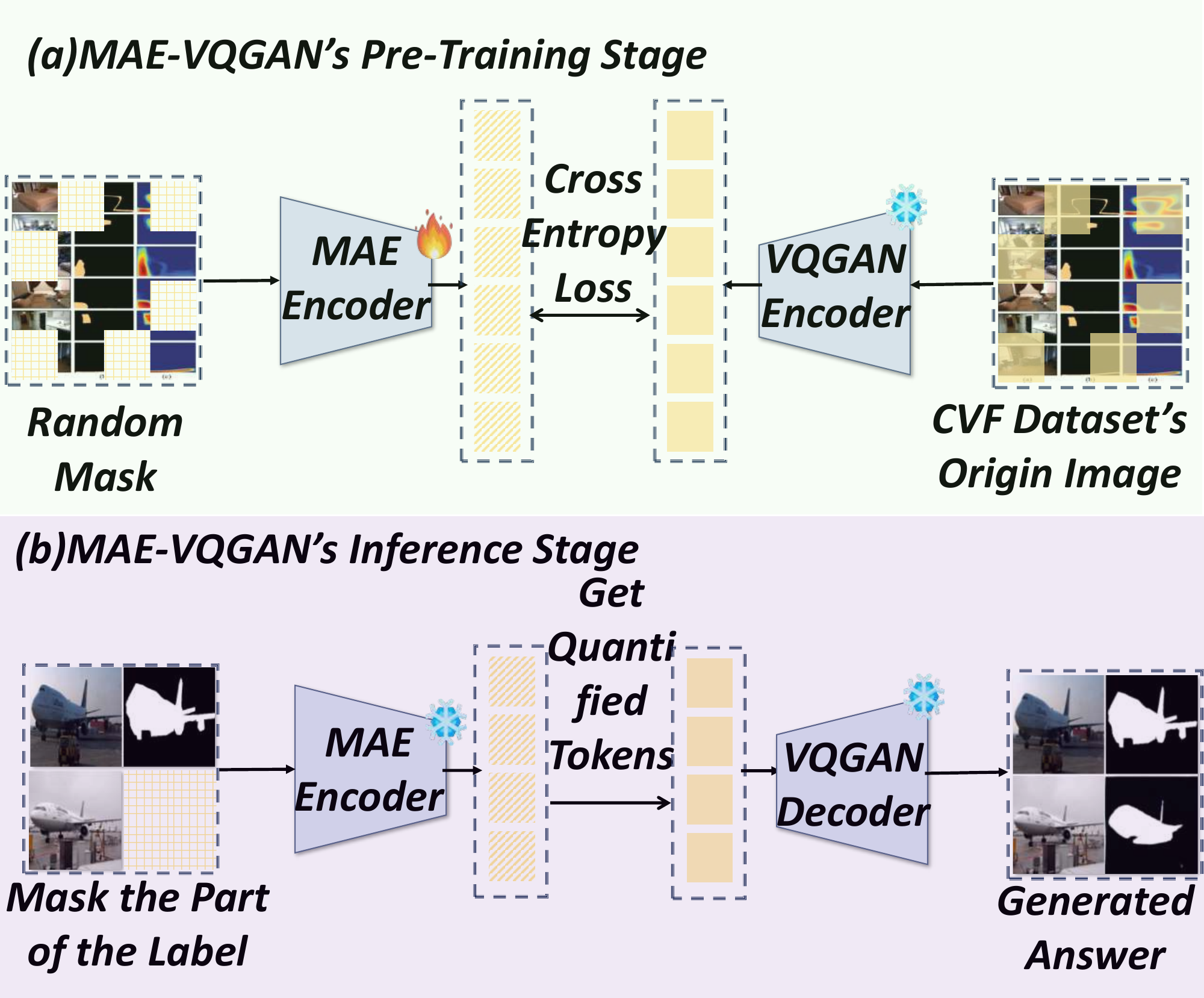}
    \caption{
Introduction to MAE-VQGAN \citep{bar2022visual}:
(a) In the pre-training stage, MAE \citep{he2022masked} is trained to enhance its inference capability through a masked reconstruction task on CVF dataset.
(b) In the inference stage, the prompt pair is placed above, with the query positioned below, and both are fed into the model for generative processing.
    }
    \vspace{-1.5em}
    \label{fig:maevqgan}
\end{wrapfigure}

\section{Preliminary: MAE-VQGAN}\label{subsec:maevqgan}

As described in \Cref{fig:maevqgan}, MAE-VQGAN \citep{bar2022visual}, comprising the MAE \citep{he2022masked} and VQGAN components, serves as a backbone for VICL through an in-painting approach. Given an example and query for the current task, MAE-VQGAN is treated as a versatile model capable of solving several image-to-image tasks.

During the pre-training phase, the model is trained on a dataset CVF, where each image is constructed from multiple sub-images, proceeding the masked reconstruction task. This process fine-tune the MAE encoder to align the distances with its the VQGAN's codebook space. In the inference phase, a in-context sample is fed into the MAE encoder, and the corresponding content from the VQGAN's codebook space is obtained, which is then passed to the VQGAN decoder for generating the output.

We utilizes the pre-trained parameters of MAE-VQGAN, freezing its parameters throughout the entire process.

\clearpage

\section{Inference Time and GPU Overhead}

\begin{wraptable}{r}{0.5\textwidth}
\centering
    \vspace{-2em}
    \caption{Comparison of the inference time and GPU overhead between \modelname{} and baselines.}
    \resizebox{0.5\columnwidth}{!}{
    \begin{tabular}{l c c} 
        \toprule
        \multirow{2}{*}{\textbf{Method}} & \textbf{Inference Time} & \textbf{GPU Cost} \\
        & \textbf{(ms/query)} & \textbf{(MB/query)} \\
        \midrule
        \midrule
        MAE-VQGAN & 51.26 & 416.14 \\
        InMeMo & 54.28 & 497.50 \\
        Prompt-SelF$_{\Number=16}$ & 984.62 & 441.75 \\
        \condenser{}$_{N=1}$ & 59.17 & 565.42 \\
        \condenser{}$_{N=16}$ & 66.61 & 1021.86 \\
        \rowcolor[HTML]{F0FFFF} \modelname{}$_{\Number=1}$& 63.14  & 569.88 \\ 
        \rowcolor[HTML]{ADD8E6} \modelname{}$_{\Number=16}$& 70.40 & 1032.50 \\
        \hline
        \bottomrule
    \end{tabular}
    }
    \label{tab:time_cost_comp_others}
\end{wraptable}

As shown in \Cref{tab:time_cost_comp_others}, we compare the inference time and GPU usage of \modelname{} with 
other baselines. The time for retrieving prompt pair is not included in the inference time. All methods that require only prompt pair retrieval are categorized under the MAE-VQGAN class. It can be observed that the time overhead of our approach increases only modestly compared to other methods, with GPU usage growing at approximately 30MB per prompt pair. Therefore, \modelname{} is resource-efficient.
This further confirms the lightweight nature of the plug-in \modelname{} based on prompt fusion, which incurs only minimal additional computational and GPU overhead. The study underscores the practical feasibility of deploying this approach in real-world scenarios, offering an effective and resource-efficient solution.

\section{Analysis of Hyperparameter}

\subsection{Analysis of Hyperparameter $\sigma$} 

\begin{wrapfigure}[]{r}{0.5\textwidth}
\centering
    \vspace{-5em}
    \includegraphics[width=0.5\columnwidth]{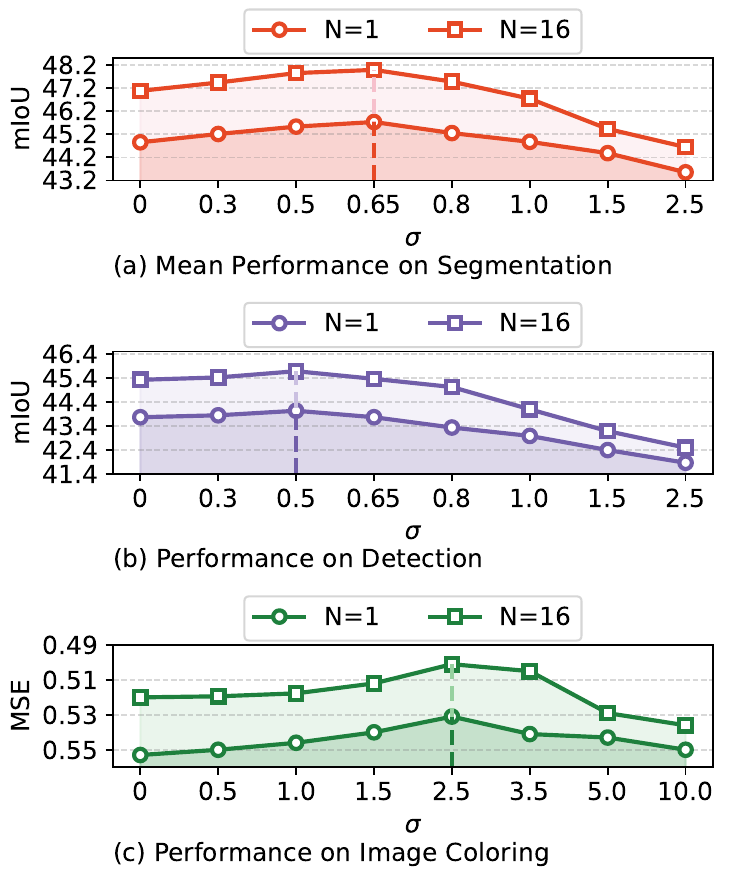}
    \vspace{-1em}
    \caption{Evaluation of \modelname{}'s performance on three tasks across varying values of $\sigma$. 
    \vspace{-3em}
    }
\label{fig:diff_sigma}
\end{wrapfigure}
The hyperparameter $\sigma$ influences the neighborhood range selected by \modelname{}. When $\sigma \to 0$, the selected neighborhood consists solely of the content of the current (h, w) token. As $\sigma \to \infty$, the selected neighborhood encompasses global information, equivalent to the standard cross-attention. As shown in \Cref{fig:diff_sigma}, extremely large or small values of $\sigma$ result in either insufficient emphasis on local information or neglect of global information. Moreover, the optimal $\sigma$ value varies across tasks. For high-level and low-level tasks, $\sigma = 0.5$ and $\sigma = 2.5$ are both reasonable choices, respectively.

\section{Experimental Analysis of Query-Conditional Sigma}

We design a straightforward query-conditioned sigma mechanism to investigate the impact of adaptive $\sigma$ for the same task. Specifically, we average the embedding dimension of the query [batchsize, patch-number, embeddim], apply a linear layer, and use a sigmoid activation to constrain the sigma value within (0,1). We report its performance on segmentation and detection tasks.

\vspace{-1em}

\begin{table*}[h]
    \centering
    \small
    \caption{Comparison of results between query-conditioned sigma and hyperparameter sigma.}
    \vspace{0.5em}
    \resizebox{0.8\columnwidth}{!}{
    \begin{tabular}{l|cccccc}
        \toprule
        \textbf{Method} & \textbf{Fold-0} & \textbf{Fold-1} & \textbf{Fold-2} & \textbf{Fold-3} & \textbf{Mean} & \textbf{Det} \\
        \midrule
        PromptHub$_{N=1}$(query-adaptive sigma) & 43.79 & 51.93 & 44.56 & 43.18 & 45.86 &  44.25 \\
        PromptHub$_{N=16}$(query-adaptive sigma) & 46.44& 52.97&45.66&46.89&47.99&45.41 \\
        PromptHub$_{N=1}$(hyperparameter sigma) & 43.26 & 50.75&43.83&42.82&45.17&44.51 \\
        PromptHub$_{N=16}$(hyperparameter sigma) & 45.93 & 53.12 & 45.44 & 46.74 & 47.81 & 45.59 \\
        \bottomrule
    \end{tabular}
    }
    \label{tab2}
    \vspace{-2mm}
\end{table*}

As shown in \Cref{tab2}, employing a simple learnable $\sigma$ within the same task yields limited improvements. This suggests that more sophisticated spatially varying priors are required, which we leave for future exploration.

\section{Exploring Multi-Objective Segmentation}

\begin{table*}[h]
    \centering
    \small
    \caption{Results of Multi-Objective Segmentation Experiments.}
    \resizebox{0.5\columnwidth}{!}{
    \begin{tabular}{l|ccccc}
        \toprule
        \textbf{Method} & \textbf{Fold-0} & \textbf{Fold-1} & \textbf{Fold-2} & \textbf{Fold-3} & \textbf{Mean} \\
        \midrule
        SupPR & 26.85 & 32.73 & 33.48 & 28.40 & 30.37 \\
        InMeMo & 28.13 & 38.31 & 37.94 & 33.08 & 34.37 \\
        PromptHub$_{N=16}$ & 38.56 & 46.54 & 45.34 & 39.23 & 42.41 \\
        \bottomrule
    \end{tabular}
    }
    \label{tab11111}
    \vspace{-2mm}
\end{table*}

We further report the numerical results on multi-object segmentation, using a subset filtered by annotations. As shown in Table \ref{tab11111}, on the complex task of multi-objective segmentation, our PromptHub model achieves an average mIoU that surpasses the strongest competitor, InMeMo, by approximately 23.4\%. This demonstrates that our approach maintains strong transferability in challenging tasks and exhibits robust generalization capability.

\section{More Visualization}

\subsection{Visualization of VICL Answer via \modelname{}}

\begin{figure*}[t]
    \centering
    \includegraphics[width=0.9\textwidth]{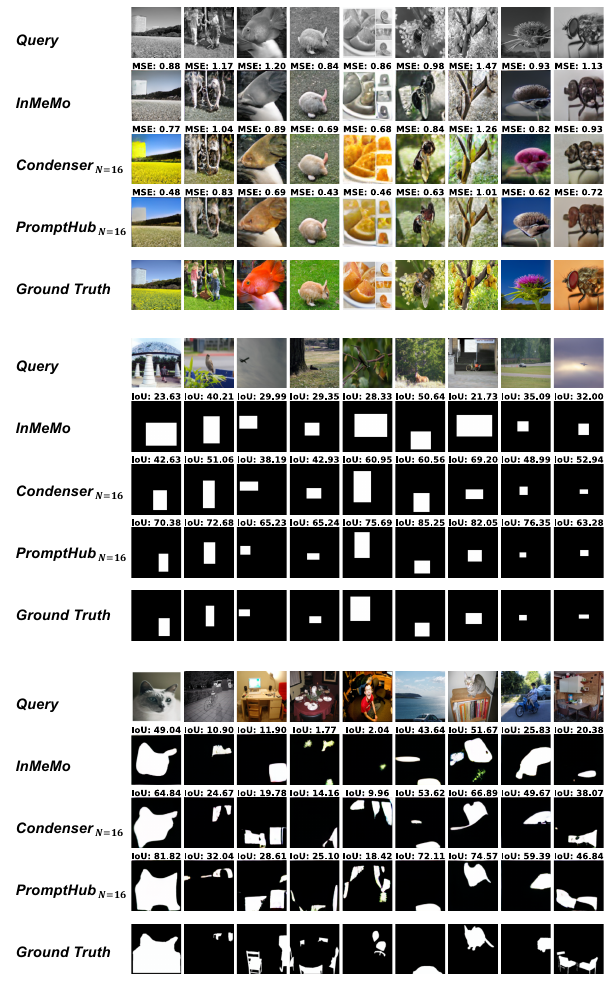}
    \caption{Comparative visualization of our method against the existing state-of-the-art method for Foreground Segmentation and Single-Object Detection and Colorization tasks. 
    }
\label{fig:exp_img}
\end{figure*}

As illustrated in \Cref{fig:exp_img}, \modelname{} consistently outperforms prior baselines across all three tasks. In particular, the segmentation and colorization results demonstrate that the predictions generated by \modelname{} exhibit smoother textures, which further substantiates the advantages of the locality-aware paradigm. Moreover, the ability of \modelname{} to strengthen multi-prompt VICL highlights its potential to drive more comprehensive progress in this domain.

\subsection{Visualization of Attention Map}

\begin{figure*}[t]
    \centering
    \includegraphics[width=0.5\textwidth]{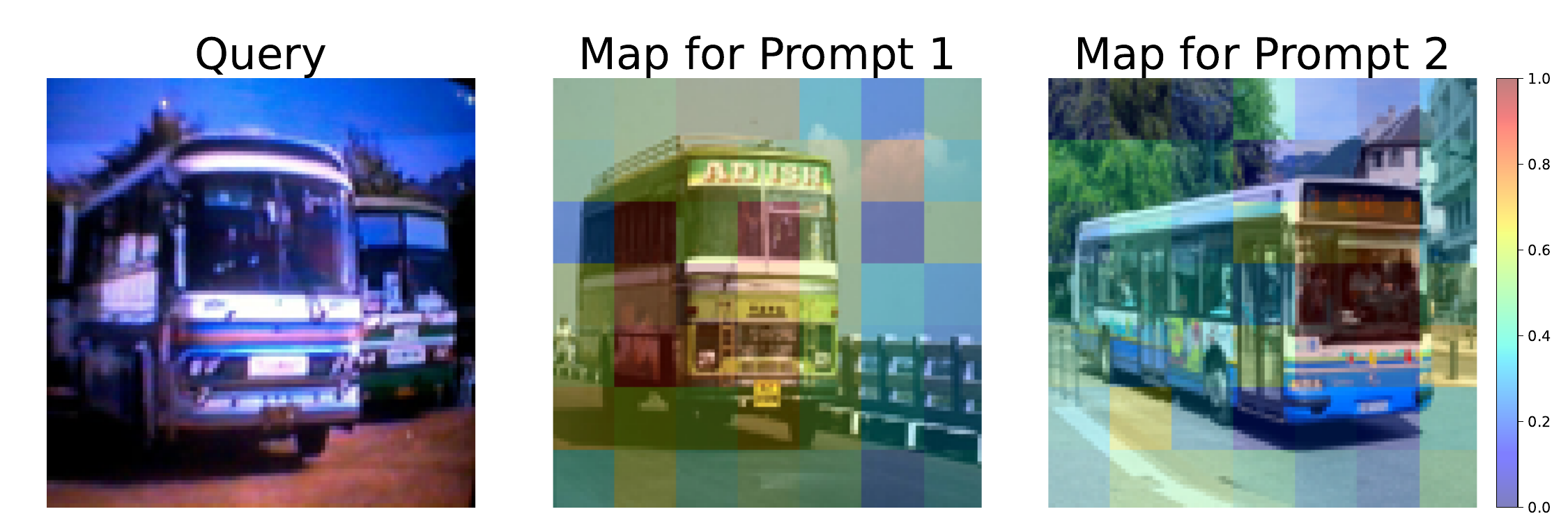}
    \includegraphics[width=0.5\textwidth]{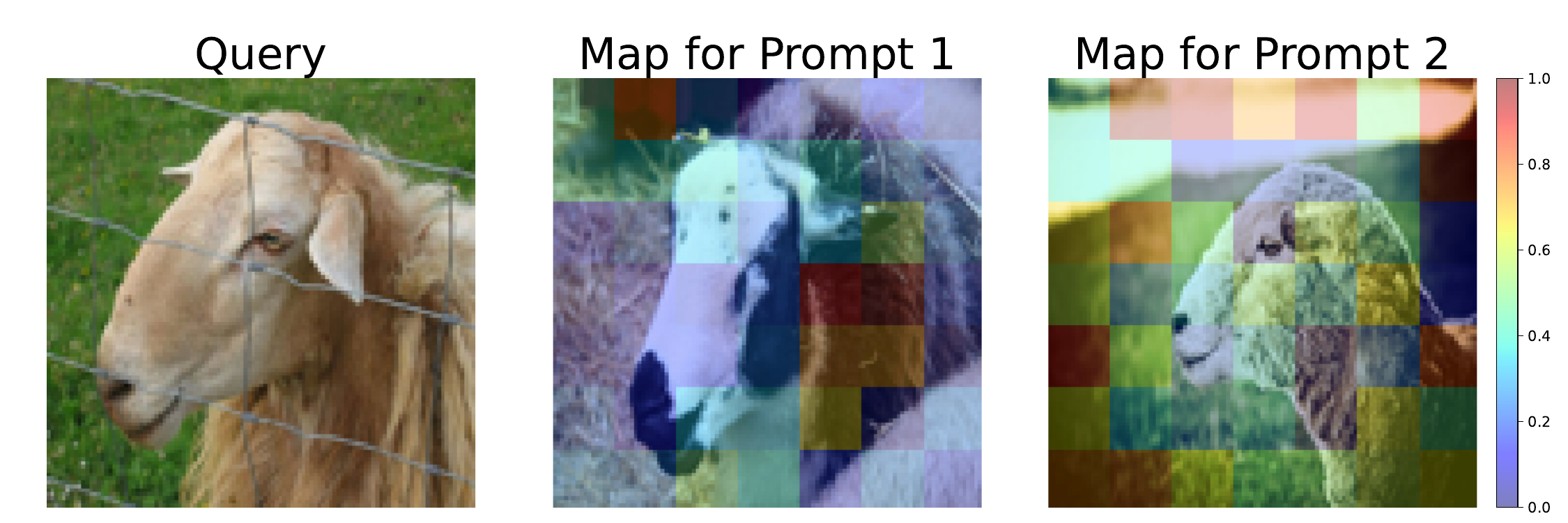}
    \includegraphics[width=0.5\textwidth]{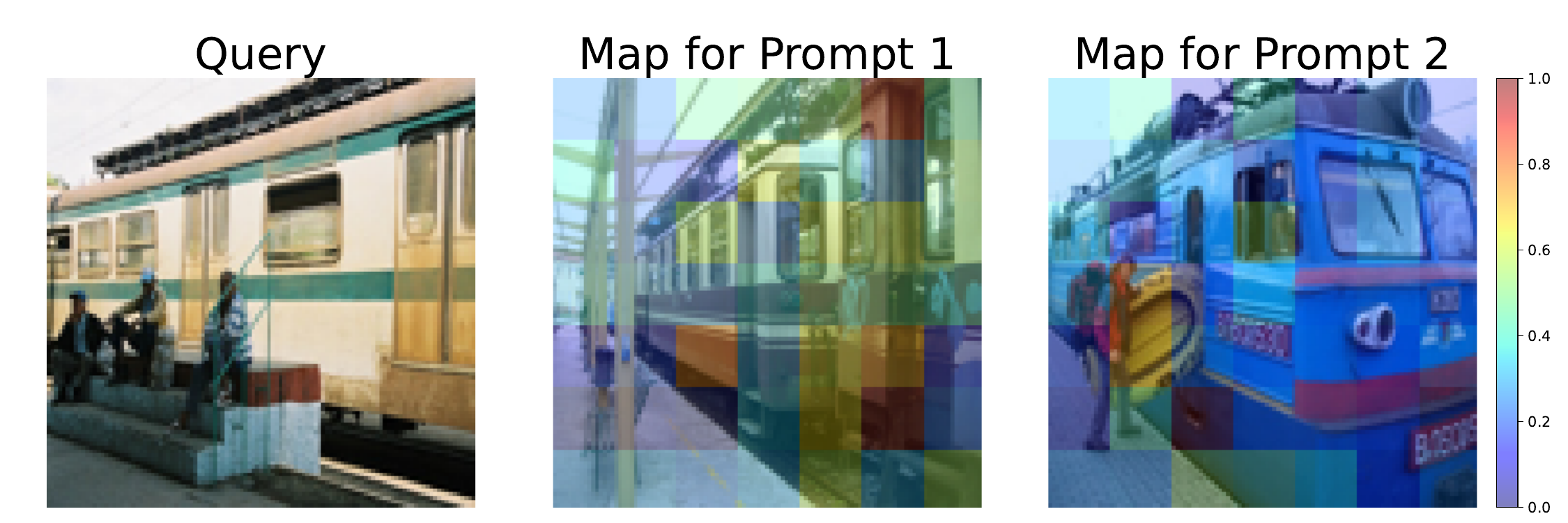}
    \includegraphics[width=0.5\textwidth]{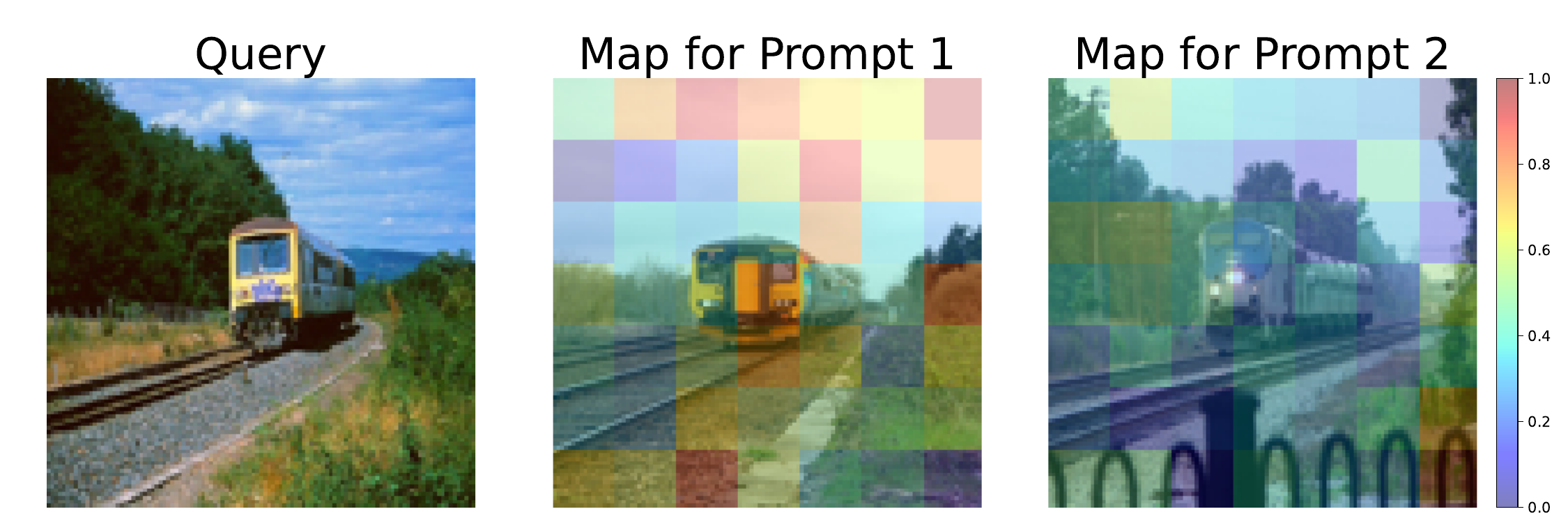}

    \caption{Visualization of attention map for $N=2$. 
    }
\label{fig:attn_vis}
\end{figure*}

As shown in \Cref{fig:attn_vis}, we visualize the attention map for prompt fusion with $N=2$, demonstrating that PromptHub effectively focuses on regions corresponding to the query image. The attention score for the current patch is computed as the normalized result of its attention score from all query patches.

\end{document}